\documentclass[runningheads]{llncs}
\usepackage[T1]{fontenc}
\usepackage{amsmath}
\usepackage{newtxtext}
\usepackage[varvw]{newtxmath}

\usepackage{amsmath,amsfonts,bm}

\def\eqref#1{equation~\ref{#1}}

\def\1{\bm{1}}

\DeclareMathAlphabet{\mathsfit}{\encodingdefault}{\sfdefault}{m}{sl}
\SetMathAlphabet{\mathsfit}{bold}{\encodingdefault}{\sfdefault}{bx}{n}

\usepackage{graphicx}
\usepackage{xcolor}
\usepackage[numbers,sort&compress]{natbib}
\renewcommand{\citet}{\cite}
\usepackage[resetlabels]{multibib}
\newcites{AP}{References for the appendix}
\usepackage{hyperref}

\usepackage{color-edits}
\addauthor{zc}{red}
\addauthor{pb}{blue}
\addauthor{pf}{orange}

\usepackage{booktabs,graphicx,xcolor,colortbl}
\definecolor{resultbest}{HTML}{39998F}
\definecolor{resultsecond}{HTML}{F5DB70}
\begin{document}
\title{Better Protein Function Prediction by Modeling Survivorship Bias}
\titlerunning{Protein Function Prediction with Survivorship Bias}
\author{Zhongmou Chao\inst{1}\thanks{Equal contribution.} \and
Poompol Buathong\inst{2}\textsuperscript{\ensuremath{\star}} \and
Ekaterina Selivanovitch\inst{1} \and
Susan Daniel\inst{1} \and
Peter I. Frazier\inst{3}}
\authorrunning{Z. Chao et al.}
\institute{Smith School of Chemical and Biomolecular Engineering, Cornell University, USA\\
\email{\href{mailto:zc83@cornell.edu}{zc83@cornell.edu}, \href{mailto:es838@cornell.edu}{es838@cornell.edu}, \href{mailto:sd386@cornell.edu}{sd386@cornell.edu}}
\and
Center for Applied Mathematics, Cornell University, USA\\
\email{\href{mailto:pb482@cornell.edu}{pb482@cornell.edu}}
\and
School of Operations Research and Information Engineering, Cornell University, USA\\
\email{\href{mailto:pf98@cornell.edu}{pf98@cornell.edu}}}
\maketitle
\begin{abstract}
Predicting whether a protein sequence possesses a biochemical property of interest is important for understanding and designing protein functionality.
Modern protein sequence databases provide valuable training data for functional prediction, but they exhibit survivorship bias: natural selection favors variants that support organismal survival while eliminating many others before they can be observed, complicating classifier training. This setting motivates positive-unlabeled (PU) learning, which treats observed functional sequences as positive and unobserved sequences as unlabeled rather than nonfunctional. 
However, existing protein PU methods do not account for cross-sequence heterogeneity in evolutionary accessibility: sequences one nucleotide mutation away from a common variant arise far more readily than distant ones.
We introduce PULSE: PU Learning with Surveillance and Evolution, a framework that trains a probabilistic classifier for a target protein function using an evolution-aware observation likelihood.
When considering functionality of proteins with high genomic sequence coverage, PULSE combines a probabilistic classifier that estimates whether an amino acid sequence performs the target function with an evolutionary model that estimates the likelihood that its encoding nucleotide sequences emerge and are detected based on mutational accessibility, parental prevalence, and surveillance efficiency.
By coupling these two components, PULSE treats the absence of an evolutionarily accessible sequence as stronger evidence of non-functionality than the absence of one unlikely to arise.
Training on genomic surveillance data and using deep mutational scanning (DMS) experimental measurements as test ground truth, PULSE ranks first or second on all eleven tasks spanning five SARS-CoV-2 variants of concern, and achieves the best overall performance among all methods, including standard PU learning, one-class classification methods, and protein language models. By accounting for evolutionary observability, PULSE enables the identification of unobserved sequences predicted to perform target functions and their prioritization for viral surveillance and drug screening.

\end{abstract}
\section{Introduction}
Understanding the relationship between protein sequence and function is central to protein engineering and human life. 
In many applications, we want to predict whether an amino acid sequence has a particular property, such as catalytic activity, binding to a target, or fusing with a type of cell membrane. Accurate prediction on how sequence alters a specific function can help identify viral variants of concern missed by surveillance, prioritize therapeutic candidates, and design de novo proteins with properties on demand. Yet, observed natural protein sequences represent only a small, and biased portion of protein sequence space: they are shaped by evolutionary and experimental selection processes that systematically overrepresent functional sequences while filtering out non-functional ones. This creates survivorship bias~\citep{bermudez2020mutational,thomas2022survivor} that limits standard supervised learning approaches to protein function prediction.

This bias has two sources. First, natural evolution preferentially retains variants with properties necessary for organism survival and reproduction, causing protein databases to be dominated by sequences that survive natural selection. Second, experimental selection protocols such as biopanning systematically enrich for sequences with a desired property, such as binding affinity to a specific target, while discarding those that fail to meet a functional threshold~\citep{giordano2001biopanning,mcguire2009biopanning}. As a result, datasets for a function of interest may contain many positive examples and few verified negatives.

This structure makes protein function prediction a positive-unlabeled (PU) learning problem~\citep{liu2003building,bekker2020learning}, where observed sequences are treated as functional positives and all others are unlabeled---they may be functional but unobserved, or truly non-functional. Existing PU methods address this setting by modeling the probability that a positive example is observed. However, these methods typically assume that this probability is constant across sequences, ignoring the fact that a sequence's probability of being observed depends on its evolutionary accessibility. This assumption discards useful information: sequences that are one mutation away from highly prevalent variants are far more likely to have been observed, if functional, than sequences requiring multiple unlikely mutations.

Alternative approaches include one-class classification (OCC), which learns a boundary or density description of observed functional variants in a chosen representation space~\citep{tax2001uniform,perera2021one}, and evolutionary sequence models, including protein language models (PLMs) and alignment-based generative models, which learn sequence constraints from protein databases or multiple sequence alignments to score variant effects~\citep{meier2021language,frazer2021disease,thadani2023learning}. However, OCCs do not explicitly model how variant function and observability jointly shape the observed data. General-purpose protein sequence models learn broad evolutionary regularities~\citep{notin2023proteingym}, and their zero-shot scores may therefore capture general mutational tolerance or net evolutionary fitness rather than optimize for any individual properties. A detailed review of relevant literature is provided in Appendix~\ref{app:litreview}.
\begin{figure}[t]
    \centering
    \includegraphics[width=\linewidth]{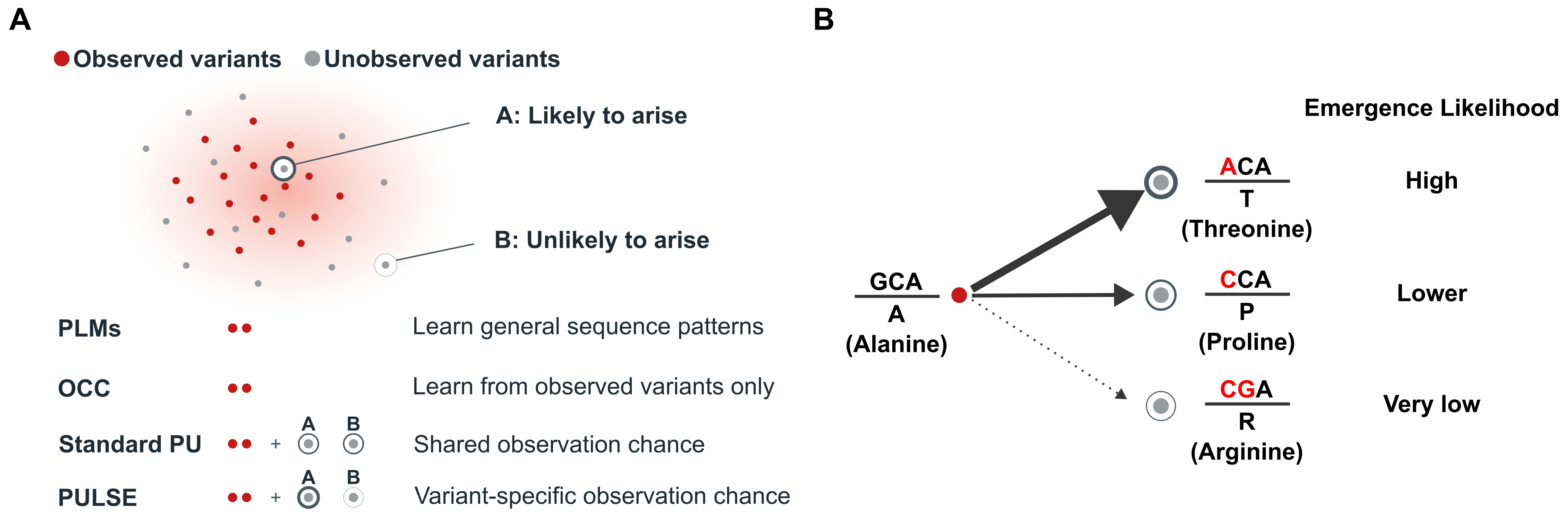}
    \caption{\textbf{(A) Overview of PULSE.} Given observed variants (red) from surveillance, PULSE models variant-specific observation probabilities for unobserved variants (gray). The larger marker for variant A than for variant B illustrates that variants can differ in their probability of being observed if both are functional. By accounting for these differences, PULSE corrects for observation bias when predicting variant function. \textbf{(B) Nucleotide-level mutations shape variant emergence.} As a simple example to demonstrate our idea, starting from an observed alanine-encoding codon, different mutational pathways (G to A via transition \textit{verus} G to C via transversion) and distances (number of nucleotide changes required, indicated by red color) can generate amino acid variants with distinct emergence likelihoods.}
    \label{fig:method_overview}
\end{figure}

To address this problem, we propose PULSE, a PU learning with surveillance and evolution framework whose core contribution is an evolution-informed, sequence-dependent observation likelihood. PULSE explicitly models the fact that functional variants are not equally likely to be observed: conditional on functionality, variants that are more mutationally accessible from prevalent circulating backgrounds have greater opportunities to arise and, consequently, to appear in surveillance data. As shown in Figure~\ref{fig:method_overview}(A), our key insight is that, functional sequences closer to those prevalent in nature are more likely to be observed than functional sequences that are evolutionarily distant, since nearby sequences are more likely to arise through feasible mutational pathways. By modeling protein evolution at the nucleotide level as shown in Figure~\ref{fig:method_overview}(B), PULSE captures how mutational accessibility and prevalence jointly determine which functional sequences are observed. This helps the classifier learn a target function from observed positives while interpreting the absence of other sequences according to their opportunities to emerge. Importantly, the observation likelihood proposed by PULSE can be used as a modular component with different sequence representations and predictive architectures, providing a plug-and-play framework for PU learning from surveillance data. The current formulation is designed for single-organism sequence data with relatively uniform surveillance coverage.

Using publicly available SARS-CoV-2 genomic sequencing data, we train PULSE to model sequence variation in the Spike protein receptor-binding domain (RBD) on multiple RBD-mediated properties and evaluate its predictions against published deep mutational scanning (DMS) measurements. We benchmark PULSE against standard PU learning, OCCs, and PLM-based approaches. PULSE ranks first or second on all 11 tasks spanning five SARS-CoV-2 variants of concern, and achieves the best overall performance among all methods.

\section{PULSE Method}
\label{sec:pulse-method}


\subsection{Data-generating process and PULSE likelihood}
\label{sec:data-generating}

We model how functional protein sequences emerge through mutation and are observed through surveillance. Let $\mathcal{A}$ denote the 20 natural amino acids and $\mathcal{X}$ the
amino acid sequences of length $L$. For $x\in\mathcal{X}$, let
$A(x)\in\{0,1\}$ indicate whether $x$ exhibits the property of interest. We model the probability of a sequence $x$ exhibiting the property of interest by a probabilistic classifier parameterized by $\theta$, i.e.
$
P(A(x)=1)=p_a(x;\theta)
$ and we aim to estimate this $\theta$.
PULSE is classifier-agnostic and supports any continuously differentiable
architecture.

Nucleotide sequences are translated into amino acid sequences through the
genetic code. Let $\mathcal{N}$ be the nucleotide alphabet and
$\mathcal{Y}$ the valid nucleotide sequences of length $3L$ that translate
into $\mathcal{X}$. We define the translation map
$B:\mathcal{Y}\rightarrow\mathcal{X}$ and the encodings of each $x$ by
$
\mathcal{Y}(x)=\{y\in\mathcal{Y}:B(y)=x\}.
$

\emph{Emergence} denotes mutation-based generation, before subsequent
viability and selection. Let $E(y)\in\{0,1\}$ indicate whether $y$
emerges, and define
$
p_e(y;\beta)=P(E(y)=1),
$
where $\beta$ contains the emergence-model parameters specified in
Section~\ref{sec:emergence}.

Let $O_\mathcal{Y}(y)\in\{0,1\}$ indicate whether $y$ is observed, and
let $D_\mathcal{Y}\subseteq\mathcal{Y}$ be the observed nucleotide
sequences. A sequence can be observed only if it emerges, encodes a
functional protein, and is detected with unknown probability
$p_o\in(0,1)$:
\[
P\!\left(O_\mathcal{Y}(y)=1\mid E(y),A(B(y))\right)
=
p_o E(y)A(B(y)).
\]

Similarly, let $O_\mathcal{X}(x)$ indicate whether amino acid sequence $x$
is observed. The observed set is
$D_\mathcal{X}=\{B(y):y\in D_\mathcal{Y}\}$, with
$
O_\mathcal{X}(x)
=
\mathbf{1}\!\left\{
\exists\,y\in\mathcal{Y}(x):O_\mathcal{Y}(y)=1
\right\}.
$

For a functional amino acid sequence $x$, define its observation
probability as
\[
\omega(x;p_o,\beta)
=
1-\prod_{y\in\mathcal{Y}(x)}
\left(1-p_o p_e(y;\beta)\right).
\]
This accounts for all encodings of $x$: it remains unobserved only when
none both emerges and is detected. Combining functionality, emergence,
and surveillance gives
\[
P(O_\mathcal{X}(x)=1)
=
p_a(x;\theta)\,\omega(x;p_o,\beta).
\]

Let $D_\mathcal{X}'=\mathcal{X}\setminus D_\mathcal{X}$ denote unobserved
amino acid sequences. The PULSE log-likelihood is
\begin{equation}
\ell_n(\theta,p_o,\beta;D_\mathcal{X})
=
\sum_{x\in D_\mathcal{X}}
\log\!\left[
p_a(x;\theta)\,\omega(x;p_o,\beta)
\right]
+
\sum_{x'\in D_\mathcal{X}'}
\log\!\left[
1-p_a(x';\theta)\,\omega(x';p_o,\beta)
\right].
\label{eq:pulse-likelihood}
\end{equation}

Exact evaluation is computationally challenging for large $L$ because
$D_\mathcal{X}'$ and the encoding sets $\mathcal{Y}(x)$ can be
exponentially large. Section~\ref{sec:approximation} therefore restricts
the candidate sets to retain the dominant likelihood contributions, and
PULSE is trained by jointly maximizing the approximate likelihood over
the classifier and observation-model parameters.

\subsection{Comparison of PULSE with binary classification, PU-learning and PLMs}
\label{sec:comparison}

The distinction of PULSE with classic binary classification and the PU-learning baselines is how it models the observation probability of a functional sequence:
\begin{equation}
P(O_\mathcal{X}(x)=1)
=
\begin{cases}
p_a(x;\theta),
    & \text{classical binary classification},\\[3pt]
q\,p_a(x;\theta),
    & \text{constant-labeling PU},\\[3pt]
p_a(x;\theta)\,\omega(x;p_o,\alpha),
    & \text{PULSE}.
\end{cases}
\label{eq:observation-model-comparison}
\end{equation}

\paragraph{\textbf{Classical binary classification (CBC).}}
CBC treats observed sequences as functional and unobserved sequences as
non-functional, equivalently assuming that every functional sequence is
observed. PULSE allows functional sequences to remain unobserved because
either they never emerges or surveillance misses them. If we set
$\omega(x;p_o,\alpha)=1$ for every $x$, PULSE recovers CBC.

\paragraph{\textbf{Constant-labeling PU (Const-PU).}}
Const-PU also permits unobserved functional sequences but assigns every
sequence the same labeling probability $q\in(0,1)$. PULSE replaces this
global probability with $\omega(x;p_o,\alpha)$, allowing sequences with
similar functionality probabilities to have different observation
probabilities because of vastly different mutational accessibility (for example, threonine
and arginine in Fig.~\ref{fig:method_overview}(B)). Setting
$\omega(x;p_o,\alpha)=q$ recovers Const-PU.

This baseline is a one-sample abstraction of PU learning rather than the
original likelihood of~\cite{song2021inferring}, which models separate
samples from an unselected library and a positive library obtained after
selection; PULSE instead models whether sequences appear in a single
observed dataset.

\paragraph{\textbf{PLM-based methods.}}
PLMs learn statistical patterns from observed protein sequences, and
their likelihood or variant-effect scores need not represent the
probability of a specific biochemical property. PULSE explicitly trains
$p_a(x;\theta)$ for the property of interest and models absence through
nucleotide emergence and surveillance, rather than treating the observed
distribution as a complete description of sequence space.

\subsection{Nucleotide emergence model}
\label{sec:emergence}

New nucleotide sequences may emerge through mutational pathways from
multiple observed sequences. We restrict possible ancestors to a
high-prevalence \emph{parent set}. Let $f(y)\in[0,1]$ denote the
prevalence fraction of $y$, typically estimated from its frequency in
$D_\mathcal{Y}$, and define
$
D_\mathcal{Y}^P=\{y\in D_\mathcal{Y}:f(y)\geq\tau\}
$
to be the high-prevalance parent set, where $\tau$ is a prevalence threshold.

For candidate nucleotide $y$, let $\mathcal{P}(y)=D_\mathcal{Y}^P$ be its available
parents, and let $P(y^p\rightarrow y)$ denote the probability that parent
$y^p$ mutates to $y$ in one generation (Section~\ref{sec:mutation}).

The effective number of opportunities for parent $y^p$ to produce progeny
is $C f(y^p)$, where $C$ estimates the total number of sequences.
 Let $\alpha>0$ account for additional factors affecting successful
emergence. Assuming independent opportunities, the probability that at
least one parent produces $y$ is
\begin{equation}
p_e(y;\alpha,C)
=
1-
\prod_{y^p\in\mathcal{P}(y)}
\left(1-\alpha P(y^p\rightarrow y)\right)^{C f(y^p)}.
\label{eq:emergence-exact}
\end{equation}
Thus, emergence probability accumulates across parents and increases with
their prevalence and mutational accessibility to $y$.

Because mutation probabilities and $\alpha$ are typically small, we apply
$(1-z)^c\approx e^{-zc}$:
\begin{equation}
p_e(y;\beta)
\approx
1-
\exp\!\left[
-\beta
\sum_{y^p\in\mathcal{P}(y)}
P(y^p\rightarrow y)f(y^p)
\right],
\label{eq:emergence-approx}
\end{equation}
where $\beta=\alpha C$ combines the global scaling quantities and is
estimated directly.

We use the same emergence model for observed and unobserved candidates;
Section~\ref{sec:approximation} describes their construction and
incorporation into the approximate PULSE likelihood.

\subsection{Mutation rate model}
\label{sec:mutation}

We model mutation using a single-site substitution matrix $M$ on the RNA
alphabet $\mathcal{N}=\{A,G,C,U\}$, where $M_{ab}$ is the probability
that nucleotide $a$ becomes $b$ in one generation. We define
\[
M_{ab}
=
\begin{cases}
r_{ab}, & a\neq b,\\[3pt]
1-\displaystyle\sum_{c\neq a}r_{ac}, & a=b,
\end{cases}
\]
where $r_{ab}$ is the per-generation rate of substitution
$a\rightarrow b$. These rates may differ across ordered nucleotide pairs,
capturing directional mutational biases. Because mutation rates are
small (for RNA virus, it is in the order of $10^{-6}$ to $10^{-4}$ per nucleotide per cycle~\cite{sanjuan2010viral}), we approximate the corresponding one-generation probabilities by
these rates.
Assuming independence across nucleotide positions, the probability that
parent $y$ produces candidate $y'$ in one generation is
\begin{equation}
P(y\rightarrow y')
=
\prod_{j=1}^{3L}M_{y_j,y'_j},
\label{eq:mutation-probability}
\end{equation}
where $y_j$ and $y'_j$ are the nucleotides at position $j$, and $3L$ is
the sequence length in nucleotides.

\subsection{PULSE likelihood approximation}
\label{sec:approximation}

Exact evaluation of Equation~(\ref{eq:pulse-likelihood}) is intractable for
large $L$ because the unobserved sequence space and nucleotide encoding
sets can be exponentially large. We therefore restrict the likelihood to
candidates within local neighborhoods of the parent set
$D_\mathcal{Y}^P$, focusing inference on local variation while limiting
enumeration cost.

Let $d_H(y,y')$ be the Hamming distance between nucleotide sequences,
with $d_H(y,\mathcal{S})=\min_{y'\in\mathcal{S}}d_H(y,y')$ for a set
$\mathcal{S}$. The observed candidate set excludes parents and contains
observed sequences within distance $r_O$:
\[
D_\mathcal{Y}^O
=
\left\{
y\in D_\mathcal{Y}:
0<d_H(y,D_\mathcal{Y}^P)\leq r_O
\right\}.
\]
Candidates near multiple parents are included once, but all parents
contribute to their emergence probabilities through
Equation~\ref{eq:emergence-approx}.

The unobserved candidate set contains unobserved sequences within
distance $r_U$:
\[
D_\mathcal{Y}^U
=
\left\{
y\in\mathcal{Y}\setminus D_\mathcal{Y}:
d_H(y,D_\mathcal{Y}^P)\leq r_U
\right\}.
\]
The radii $r_O$ and $r_U$, specified in the model-choice section, control
the scope and computational cost of the observed and unobserved
likelihood contributions. By construction, $D_\mathcal{Y}^U$ is disjoint
from both the parent and observed candidate sets.

The corresponding amino acid sets are
\[
D_\mathcal{X}^O=\{B(y):y\in D_\mathcal{Y}^O\},
\qquad
D_\mathcal{X}^U=\{B(y):y\in D_\mathcal{Y}^U\},
\]
which supply the observed and unobserved likelihood terms, respectively.

For each amino acid candidate $x$, we restrict the encodings used in the
observation probability to
$
\widehat{\mathcal{Y}}(x)
=
\mathcal{Y}(x)\cap
\left(D_\mathcal{Y}^O\cup D_\mathcal{Y}^U\right),
$
giving
\[
\widehat{\omega}(x;p_o,\beta)
=
1-
\prod_{y\in\widehat{\mathcal{Y}}(x)}
\left(1-p_o p_e(y;\beta)\right),
\]
where $p_e(y;\beta)$ is calculated from the parent set using
Equation.~\ref{eq:emergence-approx} for both observed and unobserved
candidates.

Substituting these practical sequence and encoding sets gives the
approximate PULSE log-likelihood
\begin{equation}
\begin{aligned}
\widehat{\ell}_n(\theta,p_o,\beta)
={}&
\sum_{x\in D_\mathcal{X}^O}
\log\!\left[
p_a(x;\theta)\,\widehat{\omega}(x;p_o,\beta)
\right]+
\sum_{x'\in D_\mathcal{X}^U}
\log\!\left[
1-p_a(x';\theta)\,\widehat{\omega}(x';p_o,\beta)
\right].
\end{aligned}
\label{eq:pulse-approx-likelihood}
\end{equation}
We jointly estimate the classifier and observation-model parameters
through
\begin{equation}
(\theta^*,p_o^*,\beta^*)
\in
\underset{\substack{\theta,\;p_o\in(0,1),\;\beta>0}}
{\operatorname{arg\,min}}
\left[-\widehat{\ell}_n(\theta,p_o,\beta)\right].
\label{eq:pulse-optimization}
\end{equation}
\section{Numerical Experiments}
We evaluate PULSE using SARS-CoV-2 viral genomic surveillance data from
GISAID~\citep{shu2017gisaid}, where survivorship bias and mutational
accessibility are especially informative. We train PULSE to model sequence
variation in the Spike protein RBD and evaluate its predictions against
published DMS measurements of ACE2 binding, yeast surface expression, and Spike protein-mediated cell entry categorized in ViroGym benchmarks~\cite{zhou2026virogym}. ACE2 binding provides the most direct
assessment of the RBD's canonical molecular function---receptor recognition,
whereas RBD surface expression serves as a proxy for variant folding and
stability in the yeast-display system. Cell entry provides a more integrated
downstream phenotype; for consistency with the other tasks, we restrict this
evaluation to variants from RBD-mutagenized libraries.

\subsection{Problem background and datasets}
\textbf{Background:} Many SARS-CoV-2 phenotypes relevant to infection and
immune evasion involve the Spike RBD, which mediates ACE2 recognition and
contains major neutralizing-antibody epitopes. DMS systematically measures how
thousands of mutations affect these properties: Starr et
al.~\cite{starr2020deep} mapped RBD expression and ACE2 binding, and Dadonaite
et al.~\cite{dadonaite2024spike} measured mutation effects on pseudovirus cell entry. These quantitative genotype--phenotype maps provide independent benchmarks for evaluating whether models trained on observed sequences recover experimentally measured functions. However, most DMS studies measure single substitutions in a fixed genetic background, whereas circulating variants often contain multiple, potentially interacting mutations. As multi-mutant libraries grow
combinatorially, exhaustive experimental characterization is impractical and ML models can complement DMS by predicting unmeasured variants. In this work, we evaluate on eleven DMS benchmarks---five ACE2 binding, five RBD expression, and one cell entry task---spanning five RBD backgrounds: Wuhan-Hu-1, Alpha, Beta, Eta, and Omicron XBB.1.5.
\textbf{Datasets:} We analyze SARS-CoV-2 spike sequences deposited in GISAID
before 2026. For each RBD background, sequences are aligned to the corresponding RBD reference, and the aligned RBD-encoding nucleotide sequences are extracted. Entries that are incomplete, cannot be reliably translated or aligned, or contain more than $10\%$ gaps or ambiguous residues within the RBD are excluded. Unique RBD nucleotide variants are then counted to quantify their prevalence among approximately 18 million GISAID entries.

\subsection{PULSE: Model choices}
\label{subsec:modelchoices}
\textbf{Parent, observed and unobserved set constructions:} We instantiate the
sequence sets of Section~\ref{sec:approximation}
 as follows. The parent set
$D^P_\mathcal{Y} = \{y \in D_\mathcal{Y} : f(y) \ge \tau\}$ uses a prevalence threshold of
$\tau = 10^{-3}$, i.e., parents are surveillance sequences that account for at
least $0.1\%$ of all reported counts; these sequences are both reliably sampled
and epidemiologically successful enough to seed new variation. We set
$r_O = 1$, so the observed candidate set $D^O_\mathcal{Y}$ contains every observed
sequence within one nucleotide substitution of at least one parent. Restricting
$r_O$ to a single substitution is deliberate: observed sequences lying further
from every parent appear only at extremely low frequencies. These distant
sequences more plausibly descend from lineages that surveillance missed, so we
exclude them rather than assign them near-zero mutation probabilities. For the
unobserved candidates we set $r_U = 3$, which covers the local neighborhood
around each parent while keeping $D^U_\mathcal{Y}$, whose size grows combinatorially with
$r_U$, computationally tractable. The amino acid candidate sets $D^O_\mathcal{X}$ and
$D^U_\mathcal{X}$ are obtained from these sets by translation.
\textbf{Amino acid representation:} Each amino acid is represented by three
standardized biochemical descriptors---hydrophobicity, molar refractivity, and
side-chain volume---so a sequence of length $L$ is encoded as a vector in
$\mathbb{R}^{3L}$.
\textbf{Classifier architecture:} We model the activity $p_a(x;\theta)$ with a
neural additive model (NAM)~\cite{agarwal2021nam}: each position has its own one-hidden-layer network
mapping the three descriptors at that position to a scalar contribution, and the
per-site contributions are summed, scaled by $1/\sqrt{L}$, and passed through a
logistic function. A detailed description is presented in ~\ref{app:nam}.
\textbf{Mutation rates:} The substitution-rate matrix $M$ follows the
directional mutational spectrum of SARS-CoV-2 estimated by~\cite{yi2021mutational}, with a genome-average rate of
$2 \times 10^{-4}$ per nucleotide per generation; the full specification
is presented in Appendix~\ref{app:mutation}.
\textbf{Optimization:} Since the sequences are long and the unobserved
candidate set is large, we optimize the approximate likelihood in Equation~\ref{eq:pulse-approx-likelihood} by
stochastic gradient descent (SGD), evaluating the observed term exactly and
estimating the unobserved term by importance sampling over $D^U_\mathcal{Y}$.

\subsection{Comparison methods and metrics}
\label{subsec:compare}
We benchmark PULSE against three families of methods: OCC, PU learning, and
PLM-based scoring, briefly described below; full details and optimization
routines are deferred to Appendix~\ref{app:opt}.

\textbf{OCC.} These methods learn from the observed
sequences only, using the same amino acid descriptor representation as PULSE.
\textit{OCSVM}~\cite{scholkopf2001estimating} learns a hyperplane that
separates the observed sequences from the origin with maximum margin.
\textit{iForest}~\cite{xu2023deep} scores sequences by their isolation depth in
an ensemble of random partitioning trees.

\textbf{PU learning.} These baselines correspond to the
observation models in Equation~\ref{eq:observation-model-comparison} of
Section~\ref{sec:comparison} and use the same NAM classifier
architecture and optimization procedure as PULSE, differing only in the
observation model. For both \textit{CBC} and \textit{Const-PU}, we weight the
two terms of the loss to account for the imbalance between the observed and
unobserved groups, normalizing the observed term by the number of observed
candidates and the unobserved term by the number of unobserved candidates.
PULSE requires no such weighting: its likelihood in
Equation~\ref{eq:pulse-approx-likelihood} already accounts for this imbalance through
the sequence-dependent observation probability $\hat{\omega}(x; p_o, \beta)$,
which weights each unobserved candidate individually according to its
mutational accessibility and surveillance outcome rather than applying a single
group-level correction. Both are trained by SGD, sampling unobserved candidates
uniformly from sequences within three nucleotide substitutions of the parent
set, matching the construction of $D^U_\mathcal{Y}$. \textit{2Step} follows the
reliable-negative heuristic: it first identifies reliable negatives from a set
of unlabeled sequences using the spy technique, and then trains a binary
classifier on the observed positives and these reliable negatives. We provide
1{,}000 unlabeled sequences, sampled uniformly from sequences within three
substitutions of the parent set, matching the construction of
$D^U_\mathcal{Y}$.; ablations with 100 and 10{,}000 unlabeled sequences are
presented in Appendix~\ref{app:ablation:2step}.


\textbf{PLM-based methods.} \textit{ESM1v}~\cite{meier2021language} scores each
variant by the zero-shot log-likelihood ratio against the corresponding
wild-type sequence under a pretrained PLM, without any
task-specific training. \textit{EVE}~\cite{frazer2021disease} trains a
variational autoencoder on a multiple sequence alignment of the protein family
and scores variants by their evolutionary index relative to the wild type.

\textbf{Metrics.} For each task, we report the Spearman correlation $\rho$
between each method's predicted score and the corresponding DMS measurement. To
summarize performance across all eleven tasks, we additionally rank the methods
within each task and report each method's average rank.

Our implementation of PULSE and all baseline methods except EVE is available
at \url{https://github.com/zhongmouchao/PULSE}. For EVE, we used the implementation from the original paper~\cite{frazer2021disease}.

\subsection{Results and discussion}
Table~\ref{tab:benchmark-results} reports the Spearman correlation $\rho$
for each method on all eleven
tasks. PULSE achieves the highest correlation on all five ACE2 binding tasks and
ties EVE for the best correlation on Beta expression, and it ranks second on
each of the remaining five tasks. PULSE is therefore the only method that ranks
first or second on every task, and its average rank of $1.50$ is the best
overall, followed by EVE ($2.41$) and CBC ($3.14$).

\begin{table*}[t]
\centering
\caption{Performance on eleven RBD prediction tasks, measured by Spearman
correlation $\rho$ (higher is better). 2Step entries report mean $\pm$
standard deviation over $100$ runs (ten unlabeled sets, ten spy assignments
each). The final row gives each method's average rank across tasks (lower is
better). Teal and yellow mark the best and second-best entries, including
ties.}
\label{tab:benchmark-results}
\setlength{\tabcolsep}{4pt}
\renewcommand{\arraystretch}{1.15}
\resizebox{\textwidth}{!}{%
\begin{tabular}{@{}llrrrrrrrr@{}}
\toprule
Task & Variant & iForest & OCSVM & Const-PU & CBC & 2Step & ESM1v & EVE & PULSE\\
\midrule
Binding & Wuhan Hu-1 & 0.30 & 0.29 & \cellcolor{resultsecond}\textbf{0.31} & \cellcolor{resultsecond}\textbf{0.31} & $0.06 \pm 0.06$ & 0.03 & 0.25 & \cellcolor{resultbest}\textcolor{white}{\textbf{0.39}} \\
Expression & Wuhan Hu-1 & 0.30 & 0.25 & 0.33 & 0.31 & $0.02 \pm 0.04$ & 0.02 & \cellcolor{resultbest}\textcolor{white}{\textbf{0.44}} & \cellcolor{resultsecond}\textbf{0.38} \\
Binding & Omicron XBB.1.5 & 0.15 & 0.26 & 0.26 & \cellcolor{resultsecond}\textbf{0.28} & $0.12 \pm 0.04$ & 0.12 & 0.25 & \cellcolor{resultbest}\textcolor{white}{\textbf{0.38}} \\
Expression & Omicron XBB.1.5 & 0.32 & 0.26 & 0.26 & 0.26 & $0.03 \pm 0.04$ & 0.12 & \cellcolor{resultbest}\textcolor{white}{\textbf{0.47}} & \cellcolor{resultsecond}\textbf{0.38} \\
Cell entry & Omicron XBB.1.5 & 0.37 & 0.31 & 0.39 & 0.41 & $0.04 \pm 0.05$ & 0.05 & \cellcolor{resultbest}\textcolor{white}{\textbf{0.55}} & \cellcolor{resultsecond}\textbf{0.45} \\
Binding & Alpha & 0.02 & 0.21 & \cellcolor{resultsecond}\textbf{0.24} & 0.23 & $0.08 \pm 0.03$ & 0.03 & 0.23 & \cellcolor{resultbest}\textcolor{white}{\textbf{0.40}} \\
Expression & Alpha & -0.02 & 0.24 & 0.35 & 0.33 & $0.03 \pm 0.03$ & 0.03 & \cellcolor{resultbest}\textcolor{white}{\textbf{0.47}} & \cellcolor{resultsecond}\textbf{0.42} \\
Binding & Beta & -0.03 & 0.21 & 0.19 & \cellcolor{resultsecond}\textbf{0.26} & $0.09 \pm 0.02$ & 0.01 & 0.24 & \cellcolor{resultbest}\textcolor{white}{\textbf{0.43}} \\
Expression & Beta & -0.02 & 0.25 & 0.28 & 0.36 & $0.03 \pm 0.03$ & -0.01 & \cellcolor{resultbest}\textcolor{white}{\textbf{0.41}} & \cellcolor{resultbest}\textcolor{white}{\textbf{0.41}} \\
Binding & Eta & 0.03 & 0.23 & 0.27 & \cellcolor{resultsecond}\textbf{0.28} & $0.09 \pm 0.04$ & 0.00 & \cellcolor{resultsecond}\textbf{0.28} & \cellcolor{resultbest}\textcolor{white}{\textbf{0.42}} \\
Expression & Eta & -0.03 & 0.24 & 0.34 & 0.36 & $0.01 \pm 0.03$ & 0.00 & \cellcolor{resultbest}\textcolor{white}{\textbf{0.46}} & \cellcolor{resultsecond}\textbf{0.42} \\
\midrule
\multicolumn{2}{l}{\textbf{Average rank $\downarrow$}} & 6.36 & 4.95 & 3.64 & \textbf{3.14} & 6.77 & 7.23 & \cellcolor{resultsecond}\textbf{2.41} & \cellcolor{resultbest}\textcolor{white}{\textbf{1.50}} \\
\bottomrule
\end{tabular}%
}
\end{table*}

The results clearly suggest that, by accounting for evolutionary accessibility,
PULSE can distinguish unobserved variants with different observation
likelihoods and learn from this distinction to better infer variant function
from their absence. In contrast, CBC assumes all unobserved variants near the
observed set are non-functional, while Const-PU assigns every unobserved
variant the same probability of being observed conditional on function,
regardless of its mutational distance from prevalent sequences. For 2Step, restricting the unlabeled sequences to within a few point mutations of the parent sequences results in a highly clustered training set. Under these conditions, its spy-based procedure often identifies very few reliable negatives. This leaves the final classifier with severely imbalanced training data, leading to poor predictive performance. 
OCSVM and iForest learn only from the density of observed functional
sequences in surveillance data, making these one-class methods sensitive to
bias in the training data and poorly suited to scoring unobserved DMS variants.

Notably, PULSE outperforms EVE on all five ACE2 binding tasks, which most
directly measure the RBD's specific receptor-recognition function, whereas
EVE's advantage is concentrated in expression and cell entry. This pattern
suggests that PULSE's surveillance signal and evolution-informed observation
model are particularly effective at capturing well-established, phenotype-specific
sequence variation underlying ACE2 binding affinity. By contrast, EVE's MSA-based
representation of broad evolutionary constraints may better predict general
protein functionality, as reflected by folding-related expression and the more
integrated cell-entry phenotype. ESM1v's weak performance further suggests
that, without aggregating homologous proteins via MSA, general PLM likelihoods
capture broad evolutionary plausibility but poorly resolve RBD-specific
functional variation.
\section{Conclusion}
We introduced PULSE, an evolution-informed positive–unlabeled framework for predicting protein functions that are critical to organism survival. PULSE explicitly models survivorship bias in protein sequence data by embedding nucleotide-level mutation and natural selection into a sequence-dependent observation model, yielding a biologically grounded likelihood for amino-acid sequences. Because exact likelihood computation is intractable over the full sequence space, we develop an efficient approximation that focuses on biologically plausible nucleotide-derived variants. We evaluate PULSE on eleven SARS-CoV-2 RBD prediction tasks, where it outperforms existing methods in well-surveilled viral settings and highlights its strengths of extending survivorship-aware modeling to protein fitness prediction across multiple organisms.

PULSE also leaves room for further development. Extending PULSE to settings with heterogeneous observability and to datasets that lack prevalence information would allow survivorship bias to be modeled in more general evolutionary contexts. In addition, we anticipate that integrating transformer-based architectures could further improve performance, representing an important direction for future work.

This work has the potential for significant positive impacts across multiple domains. In drug discovery, PULSE’s ability to predict functional protein variants could accelerate the identification of therapeutic targets and the design of more effective biologics. In biomaterials design, the method could guide the engineering of proteins with desired properties for industrial applications. More broadly, by providing a principled approach to modeling survivorship bias in protein databases, this work advances scientific understanding of biochemistry and biology. We also recognize potential negative impacts if the method is misused. The ability to predict functional protein and peptide sequences could be exploited to design sequences that are dangerous to human health. We emphasize that this work is intended for beneficial applications in medicine, materials science, and basic research, and that responsible use requires careful consideration of biosecurity implications and adherence to ethical guidelines.
\section*{Acknowledgements}
ZC and SD acknowledge the Smith Fellowship for Postdoctoral Innovation from Cornell University. PF and PB were partially supported by the AI Research Institutes program, supported by the NSF and the Intel Corporation under NSF Award DMR-2433348. PB gratefully acknowledges financial support from the Development and Promotion of Science and Technology Talents Project (DPST) scholarship program of the Institute for the Promotion of Teaching Science and Technology (IPST), Ministry of Education, Thailand. ES was supported by the National Institutes of Health Ruth L. Kirschstein National Research Service Award F32 fellowship 5F32AI183691.
\newpage
\bibliographystyle{splncs04}
\bibliography{ref}
\clearpage
\appendix
\renewcommand{\thefigure}{\thesection\arabic{figure}}
\renewcommand{\thetable}{\thesection\arabic{table}}

\section{Literature review}
\label{app:litreview}
In this section, we review existing methods relevant to our PULSE framework. We first discuss general approaches, including PU learning, one-class classification (OCC), and protein language model (PLM)-based methods, and then highlight specific studies that directly address protein applications, which are most relevant to our work.

\textbf{Positive-unlabeled learning:} 
Typically, PU learning methods involve two primary steps: (1) identifying some unlabeled data as reliable negatives and (2) training a final classifier model using the positive data and reliable negatives. Examples of such methods include Spy-EM~\citepAP{APliu2002partially} and Roc-SVM~\citepAP{APli2010negative}. Alternatively, some PU learning methods treat unlabeled data as negative but assign greater importance to positive data by penalizing incorrect predictions of positive instances. Examples include biased-SVM~\citepAP{APliu2003building} and weighted logistic regression~\citepAP{APlee2003learning}. A particularly relevant PU learning method for our study is the PU learning for protein design (Protein-PU) framework~\citepAP{APsong2021inferring}, which fits a logistic regression model using positive and unlabeled data through a custom loss function that incorporates prior knowledge about the distribution of labeled data.

\textbf{One-class classification:}
OCC methods can be divided into One-class Support Vector Machine (OSVM)-based and non-OSVM-based approaches~\citepAP{APkhan2014one}. Pioneer OSVM-based methods build a smallest hyper-sphere that encloses positive samples (SVDD)~\citepAP{APtax1999data,APtax1999support,APtax2001uniform} or a hyper-plane that separates positive data from the origin (OC-SVM)~\citepAP{APscholkopf2001estimating}. Recent advances in OSVM-based methods have used neural network for feature extraction and apply traditional OSVM approaches over the extracted features~\citepAP{APerfani2016high,APghafoori2020deep}. Examples of non-OSVM-based methods includes the ones using neural network models~\citepAP{APmanevitz2001one,APskabar2003single,APchalapathy2018anomaly}, decision trees~\citepAP{APliu2008isolation,APdesir2012random,APxu2023deep}, nearest neighbors~\citepAP{APmunroe2005multi} and Bayesian classifiers~\citepAP{APwang2003one}. OCC framework has been tailored to protein-related applications. For example,~\citepAP{APmei2015novel} considered a problem of prediciting protein-protein interaction and proposed to use OSVM-based method to sample negative data first and then use the two-class SVM as a final classifier.~\citepAP{APyousef2015novel} proposed to use SVDD together with physicochemical property-based representations of proteins to classify genes with diseases of interest.

\textbf{Protein classification using protein language models:}
Recent methods for protein classification leverage deep generative models trained on multiple sequence alignments to capture amino acid distributions and evolutionary conservation. For example, zero-shot prediction via the protein language model ESM-1v~\citepAP{APmeier2021language} that computes the fitness likelihood of a queried sequence with respect to a wild type sequence. The Evolutionary Model of Variant Effect (EVE)~\citepAP{APfrazer2021disease} predicts pathogenicity by training a variational autoencoder (VAE) on MSA-derived sequences of a human protein of interest. The VAE estimates the relative likelihood of each single amino acid variant compared to the wild type, producing evolutionary indices. These indices are then used to fit a two-component Gaussian mixture model that outputs pathogenicity probabilities. Another example is EVEscape~\citepAP{APthadani2023learning}, which extends the EVE framework by combining information from evolutionary scores from EVE with protein structural and chemical information and using logistic functions to predict the likelihood of immune escape in viral variants.

\section{NAM: Network architecture}
\label{app:nam}

We model the activity $p_a(x;\theta)$
with a neural additive model~\citep{agarwal2021nam}, a neural
generalization of generalized additive models~\citep{hastie1986gam} in
which each additive component is a small neural network rather than a
fixed smooth function. In our case the additive components are sequence
positions: each position $i$ has its own one-hidden-layer network
$f_i:\mathbb{R}^3\rightarrow\mathbb{R}$ with hidden width $4$ and ReLU
activation, mapping the three descriptors of the amino acid at that
position to a scalar contribution $f_i(x_i)$. The per-site contributions
are summed, scaled by $1/\sqrt{L}$, and passed through a logistic
function. Because every position has its own weights, the model learns
position-specific chemical effects, while the additive structure keeps
each site's contribution separately identifiable, so the effect of any
single substitution can be read off directly. Logistic regression is
recovered as the special case in which every $f_i$ is linear. All
per-site output weights are initialized to zero, so $p_a(x;\theta)=0.5$
for every peptide at initialization and no initial ranking is built in.
\section{Mutation rates}
\label{app:mutation}

We model mutation using a single-site substitution matrix $M$ on the RNA
nucleotide alphabet $\mathcal{N} = \{A, G, C, U\}$, as defined in Section~\ref{sec:mutation}.
For SARS-CoV-2, we set the per-generation substitution rates $r_{ab}$
proportional to the directional mutational spectrum estimated by Yi et
al.~\citeAP{APyi2021mutational} from 351{,}525 pandemic genomes, which is dominated
by $C \to U$ and $G \to U$ changes. Table~\ref{tab:mutation-spectrum} lists
the weight $w_{ab}$ assigned to each of the twelve substitution pathways.

\begin{table}[h!]
\centering
\caption{Directional substitution weights for the SARS-CoV-2 mutation model,
proportional to the mutational spectrum of Yi et al.~\cite{yi2021mutational}.
The weights sum to $100$.}
\label{tab:mutation-spectrum}
\begin{tabular}{@{}llr@{}}
\toprule
Pathway $a \to b$ & Type & Weight $w_{ab}$ \\
\midrule
$C \to U$ & Transition   & 46.5 \\
$G \to U$ & Transversion & 18.2 \\
$A \to G$ & Transition   &  9.4 \\
$U \to C$ & Transition   &  9.4 \\
$G \to A$ & Transition   &  7.4 \\
$A \to C$ & Transversion &  1.3 \\
$A \to U$ & Transversion &  1.3 \\
$C \to A$ & Transversion &  1.3 \\
$C \to G$ & Transversion &  1.3 \\
$G \to C$ & Transversion &  1.3 \\
$U \to A$ & Transversion &  1.3 \\
$U \to G$ & Transversion &  1.3 \\
\bottomrule
\end{tabular}
\end{table}

The per-generation rates are then
\begin{equation}
r_{ab} = \frac{4\mu\, w_{ab}}{\sum_{a',b'} w_{a'b'}},
\qquad a \neq b,
\end{equation}
so that all pathway ratios match the observed spectrum while the genome-average
per-nucleotide rate equals $\mu = 2 \times 10^{-4}$ per generation. The
diagonal entries of $M$ are set to $M_{aa} = 1 - \sum_{c \neq a} r_{ac}$, and
multi-site mutation probabilities factorize across positions as in Equation~\ref{eq:mutation-probability}.


\section{Details on baselines and optimization}
\label{app:opt}

This appendix collects the training and optimization details for our PULSE method and baselines in
Section~\ref{subsec:compare}, presented one by one. Methods in the PU family
(PULSE, Const-PU, CBC, and 2Step) share the same NAM classifier architecture and the same amino acid descriptor representation described in
Section~\ref{subsec:modelchoices} , and differ only
in their loss functions; we therefore describe their optimizers in full. The
one-class methods (OCSVM, iForest) and the protein language model methods
(ESM1v, EVE) do not share this training loop, and we summarize their fitting
procedures separately.

\subsection{PULSE}
\label{app:optimization:pulse}
PULSE corresponds to the last case of
Equation~\ref{eq:observation-model-comparison},
$P(O_X(x)=1)= p_a(x;\theta)\,\omega(x;p_o,\alpha)$, where each peptide is observed only if at least one of its nucleotide encodings both emerges from the parent set and is observed. The observed part of the likelihood, the exact sum of its log contributions over all observed peptides, is evaluated at every step; this set contains roughly $45{,}000$ unique amino-acid sequences within one nucleotide substitution of the parent set for each of the five variants. The unobserved part is a finite sum over all peptides reachable from the parent set and is too large to enumerate, so at each step we redraw $128$ unobserved peptides from a proposal over the same three-substitution neighborhood used to construct $D^U_Y$ with $r_U=3$:a parent is sampled in proportion to its prevalence, and the number of
mutations it accumulates is drawn from a Poisson distribution whose mean
$\lambda = 3L\mu$ is the expected number of substitutions per
replication of the full nucleotide sequence --- the per-site
per-generation mutation rate $\mu$ times the sequence length $3L$ --- truncated to $\{1,2,3\}$ and renormalized. For the RBD tasks this gives $\lambda \approx 0.12$, so roughly $94\%$ of draws carry a single substitution and the rest carry two or three, concentrating the proposal on the most probable neighborhoods while still covering multi-mutation paths. Positions and substitute nucleotides are chosen uniformly, with draws landing on an observed peptide rejected and resampled. Each draw's contribution, $\log\bigl(1-p_a(x';\theta)\,s(x')\bigr)$, is computed deterministically by exactly enumerating its reachable encodings and their emergence supports, and is weighted by the inverse proposal probability $1/q(x')$, making the stochastic gradient unbiased for the full-sum objective. All parameters---the activity network, the emergence scale, and the observation probability $p_o$---are updated jointly with AdamW (learning rate $0.01$, gradient-norm clipping at $1$, learning rate halved after $200$ steps without validation improvement), for up to $8{,}000$ steps. A fixed validation objective is evaluated every $10$ steps, training stops after $100$ consecutive checks ($1{,}000$ steps) without an improvement of at least $10^{-4}$, and the best checkpoint is restored.
\subsection{Const-PU and CBC}
\label{app:optimization:pu-cbc}

CBC corresponds to the first case of Equation~\ref{eq:observation-model-comparison}, $P(O_X(x)=1)=p_a(x;\theta)$. At each step it redraws $128$ unobserved peptides uniformly from sequences within three nucleotide substitutions of the parent set, matching the construction of $D^U_Y$ with $r_U=3$, and minimizes the mean of $\log p_a(x;\theta)$ over all observed peptides plus the mean of $\log\bigl(1-p_a(x';\theta)\bigr)$ over that draw, the weighted loss in Section~\ref{subsec:compare}. CBC is fit once on the full observed set.

Const-PU corresponds to the second case of Equation~\ref{eq:observation-model-comparison}, $P(O_X(x)=1)=q\,p_a(x;\theta)$, with one labeling probability $q\in(0,1)$ shared by every sequence. It abstracts Protein-PU~\cite{song2021inferring}: that paper fits a positive library of size $n_p$ and an unlabeled library of size $n_u$, and multiplies the classifier by the constant $n_p/(n_p+\pi n_u)$, where $\pi$ is the positive proportion in the unlabeled library. Const-PU keeps a single constant multiplier, the $q$ in Equation~\ref{eq:observation-model-comparison},, inside the one-sample observation likelihood of Section~\ref{sec:comparison}. The loss is the mean of $\log\bigl(q\,p_a(x;\theta)\bigr)$ over all observed peptides plus the mean of $\log\bigl(1-q\,p_a(x';\theta)\bigr)$ over $128$ unobserved peptides redrawn at every step from the same three-substitution neighborhood. We choose $q$ from $\{0.1,0.2,\ldots,0.9\}$ by $10$-fold cross-validation on a single split of the observed peptides, shuffled with seed $0$. Each fold holds out one tenth of those peptides as positives: $6{,}972$--$7{,}153$ for Alpha, Beta, and Eta, $7{,}082$ or $7{,}083$ for Omicron XBB.1.5 binding, and $23{,}745$--$24{,}188$ for Wuhan Hu-1 binding, Wuhan Hu-1 expression, Omicron XBB.1.5 expression, and Omicron XBB.1.5 cell entry. Alpha, Beta, and Eta each use one split for both assays. The held-out positives are scored against $128$ unobserved peptides drawn once per fold and treated as the negative class, and the selected $q$ is the one with the highest average raw ROC AUC over the ten folds. The reported model is then refit on every observed peptide at that $q$. This selects $q=0.8$ for Wuhan Hu-1 binding, Wuhan Hu-1 expression, Alpha, and Eta, and $q=0.9$ for Omicron XBB.1.5 binding, Omicron XBB.1.5 expression, Omicron XBB.1.5 cell entry, and Beta.

Both methods use Adam with PyTorch's default decay rates $(\beta_1,\beta_2)=(0.9,0.999)$, and gradients are clipped to a maximum norm of $1$. One epoch is a single gradient step on the whole training matrix, with no mini-batching over the observed peptides. The learning rate follows PyTorch's triangular cyclic schedule between $10^{-3}$ and $10^{-1}$: it rises from $10^{-3}$ to $10^{-1}$ over $50$ steps and falls back over the next $50$, so one full cycle spans $100$ epochs. Training runs for at most $2{,}000$ epochs and stops early if the loss fails to improve on the best loss so far by at least $10^{-6}$, an absolute decrease, for $100$ consecutive epochs. The retained weights are those achieving the lowest loss.





\subsection{2Step}
\label{app:optimization:2step}

2Step uses the same optimizer and training loop as Const-PU and CBC, but fits the classifier twice, and both fits minimize binary cross-entropy on a fixed training matrix. The squared-weight penalty coefficient is $0$. For the results in Table~\ref{tab:benchmark-results}, the unlabeled set contains $K=1{,}000$ peptides drawn from sequences within three nucleotide substitutions of the parent set, matching the construction of $D^U_\mathcal{Y}$. The spy step marks $20\%$ of the observed peptides, $\lfloor 0.2\,|D_\mathcal{X}^O|\rfloor$ of them, as spies and holds them out of the positive class. The first classifier is trained with the remaining observed peptides labeled positive and with the spies together with the unlabeled peptides labeled negative. Its score on each spy peptide is recorded, and the decision threshold is the lowest of those scores. An unlabeled peptide is kept as a reliable negative when its score is strictly below that threshold. The second classifier is then trained with every observed peptide labeled positive and these reliable negatives labeled negative. If the threshold admits no reliable negative, the second fit uses the observed peptides alone. Table~\ref{tab:benchmark-results} repeats this procedure over ten unlabeled sets and ten spy assignments per set, and reports the mean $\pm$ standard deviation of the Spearman correlation $\rho$ over the resulting $100$ runs.

\subsection{One-class methods}
\label{app:optimization:occ}

Both one-class baselines are implemented in
scikit-learn~\citeAP{APpedregosa2011scikit} and are fit on the same standardized
amino acid descriptor representation as PULSE, using the observed sequences
only. \textit{iForest} uses \texttt{sklearn.ensemble.IsolationForest}, and
\textit{OCSVM} uses \texttt{sklearn.svm.OneClassSVM}. Scores are given by each
method's decision function on the DMS variants.

\subsection{Protein language model methods}
\label{app:optimization:plm}

\textbf{EVE:} In EVE~\cite{frazer2021disease}, we follow the procedures described in the original paper with the implementation publised in \href{https://github.com/OATML-Markslab/EVE}{https://github.com/OATML-Markslab/EVE}. For each variant, the receptor-binding domain (RBD) sequence was queried against a local UniRef100 database using JackHMMER (HMMER~3.4, five iterations). Hits were retained if they scored $\geq 0.3$ bits per residue, aligned to $\geq 50\%$ of the reference, and contained $\leq 50\%$ gaps, following the EVE protocol. Alignments were projected onto full-length reference coordinates, with insertion columns discarded and complete UniRef100 annotations preserved. This yielded 3{,}265--3{,}284 homologous sequences per RBD across the five variants analyzed (Wuhan-Hu-1, Alpha, Beta, Eta and Omicron XBB.1.5). We then train a variational autoencoder (VAE) on the one-hot encoded MSA sequences and use it to compute an evolutionary index for each test sequence relative to the wild type. These indices are subsequently modeled with a two-component Gaussian mixture model (GMM) to predict the functional class of each sequence. 

\noindent\textbf{Zero-shot (Esm1v):} For each DMS data, we compute the fitness likelihood difference between each test sequence and
the wild type using the ESM-1v model, following Equation. (1) in~\cite{meier2021language}.

\section{Ablation study on the size of the unlabeled set for 2Step}
\label{app:ablation:2step}

The 2Step baseline first identifies reliable negatives from a set of unlabeled
sequences using the spy technique, and then trains a binary classifier on the
observed positives and these reliable negatives. Because the spy step operates
on whatever unlabeled sequences are provided, the method may be sensitive to
how large that unlabeled set is. We therefore vary the number of unlabeled
sequences $K \in \{100, 1{,}000, 10{,}000\}$, sampled uniformly from sequences
within three nucleotide substitutions of the parent set, matching the
construction of $D^U_\mathcal{Y}$; reliable negatives are then determined by the spy
technique within each sample. For each task and each $K$, we draw ten random
unlabeled sets and repeat the full 2Step procedure ten times per set, reporting
the mean $\pm$ standard deviation of the Spearman correlation $\rho$ over the
resulting $100$ runs. The main results in Table~\ref{tab:benchmark-results} use
$K = 1{,}000$.

\begin{table}[t]
\centering
\caption{Ablation over the number of unlabeled sequences $K$ provided to
2Step, measured by Spearman correlation $\rho$ (mean $\pm$ standard deviation
over $100$ runs: ten random unlabeled sets, ten runs per set). Teal marks the
best entry in each row, including ties. All values remain well below PULSE
($0.38$--$0.45$) on every task.}
\label{tab:ablation-2step}
\setlength{\tabcolsep}{4pt}
\begin{tabular}{@{}llccc@{}}
\toprule
Task & Variant & $K = 100$ & $K = 1{,}000$ & $K = 10{,}000$ \\
\midrule
Binding    & Wuhan Hu-1      & $0.06 \pm 0.06$ & $0.06 \pm 0.06$ & \cellcolor{resultbest}\textcolor{white}{\textbf{0.07}}$\,\pm\,0.06$ \\
Expression & Wuhan Hu-1      & \cellcolor{resultbest}\textcolor{white}{\textbf{0.02}}$\,\pm\,0.04$ & \cellcolor{resultbest}\textcolor{white}{\textbf{0.02}}$\,\pm\,0.04$ & \cellcolor{resultbest}\textcolor{white}{\textbf{0.02}}$\,\pm\,0.04$ \\
Binding    & Omicron XBB.1.5 & $0.10 \pm 0.05$ & \cellcolor{resultbest}\textcolor{white}{\textbf{0.12}}$\,\pm\,0.04$ & $0.07 \pm 0.07$ \\
Expression & Omicron XBB.1.5 & $0.02 \pm 0.04$ & $0.03 \pm 0.04$ & \cellcolor{resultbest}\textcolor{white}{\textbf{0.04}}$\,\pm\,0.04$ \\
Cell entry & Omicron XBB.1.5 & $0.01 \pm 0.05$ & \cellcolor{resultbest}\textcolor{white}{\textbf{0.04}}$\,\pm\,0.05$ & \cellcolor{resultbest}\textcolor{white}{\textbf{0.04}}$\,\pm\,0.04$ \\
Binding    & Alpha           & $0.05 \pm 0.04$ & \cellcolor{resultbest}\textcolor{white}{\textbf{0.08}}$\,\pm\,0.03$ & \cellcolor{resultbest}\textcolor{white}{\textbf{0.08}}$\,\pm\,0.04$ \\
Expression & Alpha           & $0.02 \pm 0.03$ & \cellcolor{resultbest}\textcolor{white}{\textbf{0.03}}$\,\pm\,0.03$ & \cellcolor{resultbest}\textcolor{white}{\textbf{0.03}}$\,\pm\,0.03$ \\
Binding    & Beta            & $0.06 \pm 0.05$ & \cellcolor{resultbest}\textcolor{white}{\textbf{0.09}}$\,\pm\,0.02$ & $0.07 \pm 0.06$ \\
Expression & Beta            & $0.01 \pm 0.03$ & $0.03 \pm 0.03$ & \cellcolor{resultbest}\textcolor{white}{\textbf{0.04}}$\,\pm\,0.05$ \\
Binding    & Eta             & $0.07 \pm 0.06$ & \cellcolor{resultbest}\textcolor{white}{\textbf{0.09}}$\,\pm\,0.04$ & $0.08 \pm 0.05$ \\
Expression & Eta             & $0.00 \pm 0.03$ & $0.01 \pm 0.03$ & \cellcolor{resultbest}\textcolor{white}{\textbf{0.02}}$\,\pm\,0.04$ \\
\bottomrule
\end{tabular}
\end{table}

Table~\ref{tab:ablation-2step} shows that 2Step's performance does not vary
monotonically with the size of the unlabeled set: on many tasks the correlation
rises and then falls as $K$ grows. This non-monotonicity is consistent with the heuristic
nature of the method: enlarging the unlabeled set gives the spy step more
candidates but also increases the chance of mislabeling unobserved functional
sequences as reliable negatives, reintroducing survivorship bias. More
importantly, all average correlations across all tasks and all values of $K$ remain
below $0.10$---far under PULSE's correlations of $0.25$--$0.55$---so no choice
of the unlabeled set size brings 2Step close to PULSE, and our conclusions are
unaffected by this design choice.
\bibliographystyleAP{splncs04}
\bibliographyAP{ref_app}
\end{document}